\documentclass[runningheads]{llncs}
\usepackage[T1]{fontenc}
\usepackage[utf8]{inputenc}
\usepackage{graphicx}
\begin{document}
\title{Functional Equivalence with NARS}
%
%

\author{Robert Johansson\inst{1, 2} \and
Patrick Hammer\inst{1,3} \and
Tony Lofthouse\inst{1}}

\authorrunning{R. Johansson et al.}
%
\institute{Department of Psychology, Stockholm University, Sweden \and
Department of Computer and Information Science, Linköping University, Linköping, Sweden \and 
Division of Robotics, Perception and Learning, KTH Royal Institute of Technology, Stockholm, Sweden
\\
\email{\{robert.johansson, patrick.hammer, tony.lofthouse\}@psychology.su.se}}
\maketitle              
\begin{abstract}
This study explores the concept of functional equivalence within the framework of the Non-Axiomatic Reasoning System (NARS), specifically through OpenNARS for Applications (ONA). Functional equivalence allows organisms to categorize and respond to varied stimuli based on their utility rather than perceptual similarity, thus enhancing cognitive efficiency and adaptability. In this study, ONA was modified to allow the derivation of functional equivalence. This paper provides practical examples of the capability of ONA to apply learned knowledge across different functional situations, demonstrating its utility in complex problem-solving and decision-making. An extended example is included, where training of ONA aimed to learn basic human-like language abilities, using a systematic procedure in relating spoken words, objects and written words. The research carried out as part of this study extends the understanding of functional equivalence in AGI systems, and argues for its necessity for level of flexibility in learning and adapting necessary for human-level AGI.

\end{abstract}

\section{Introduction}

Functional equivalence refers to the grouping of dissimilar stimuli or objects into the same category based on their common functional roles in specific contexts, rather than their perceptual similarities \cite{schusterman1998functional}. For example, if an animal learns that a particular signal or object means “food is available,” it can apply this understanding to other dissimilar signals or objects that have been associated with food in the same way, treating them as functionally equivalent. This concept is crucial for understanding how animals navigate complex social and environmental cues efficiently \cite{schusterman1998functional}.

Functional equivalence is important from an adaptation perspective because it enables animals, including humans, to respond efficiently and effectively to their environments without needing to learn each new situation from scratch. A few motivations why functional equivalence is significant for adaptive behaviors include:

\begin{enumerate}

\item \textbf{Cognitive Efficiency:} Functional equivalence allows for cognitive economy. By treating different stimuli as equivalent, an animal doesn’t need to store and retrieve unique responses for every single variant of a stimulus it encounters. Instead, it can generalize responses across a class of stimuli, simplifying decision-making and reducing cognitive load.

\item \textbf{Rapid Learning and Adaptation: } This ability to generalize across different stimuli enables faster learning and adaptation to new environments. When an animal can apply learned knowledge about one item to others within the same functional class, it can quickly adapt its behavior to new but functionally similar situations. This rapid adaptation can be crucial for survival, especially in environments where conditions change frequently.

\item \textbf{Enhanced Problem-Solving: } Functional equivalence supports complex problem-solving and reasoning. By understanding that different elements can serve a similar function, animals and humans can make inferences and predictions about new situations based on previously encountered conditions. This capability is essential for planning and decision-making in uncertain contexts.

\item \textbf{Social and Communicative Competence: } In social animals, recognizing functional equivalence can aid in understanding social cues and communications. For example, different signals from peers may be understood as indicating the same underlying message or intent, such as warnings about predators or opportunities for food. This understanding supports more complex social interactions and strengthens group cohesion and cooperation.

\item \textbf{Transfer of Knowledge: } Functional equivalence facilitates the transfer of knowledge across different contexts or modalities. For instance, a sea lion trained to respond to visual stimuli in a certain way might apply the same response principles to auditory or tactile stimuli if it perceives them as functionally equivalent. This transferability is a hallmark of higher cognitive functions and is significant for navigating a world where sensory inputs are varied and multifaceted.

\end{enumerate}

Overall, functional equivalence is a fundamental aspect of cognitive and behavioral flexibility, enabling organisms to navigate their worlds with more agility and finesse, which is a significant advantage in the natural selection process.

In this study, we extend our previous work on \textit{generalized identity matching} that was demonstrated with OpenNARS for Applications \cite{johansson2023generalized}. This work also extends previous work in that it implements a mechanism that we previously suggested to be named \textit{contingency entailment} \cite{johanssonlofthouse2023stimulus}, that we here call \textit{functional equivalence}. The rest of the paper is structured as follows. First, we provide a brief introduction to the NARS used in this paper, OpenNARS for Applications (ONA). Then, we explain how functional equivalence works with ONA in the simplest case. An extended example of training ONA to “read” (oral naming of words) and to transfer knowledge from one domain (objects) to another (words). Finally, a discussion concludes the paper.

\section{OpenNARS for Applications}

In this study, we used OpenNARS for Applications (ONA) \cite{hammer2020opennars}, a highly effective implementation of the Non-Axiomatic Reasoning System (NARS) \cite{wang2013nalbook}. ONA differentiates itself from other Non-Axiomatic Reasoning System (NARS) implementations in several key aspects, which enhance its suitability for practical applications:

\begin{enumerate}

\item \textbf{Event-Driven Control Process:} ONA incorporates an event-driven control mechanism that departs from the more probabilistic and bag-based approach used in traditional NARS implementations, such as OpenNARS \cite{lofthouse2019alann}. This shift allows ONA to prioritize processing based on the immediacy and relevance of incoming data and tasks. The event-driven approach is particularly advantageous in dynamic environments where responses to changes must be timely and context-sensitive.

\item \textbf{Separation of Sensorimotor and Semantic Inference:} Unlike other NARS models that often blend various reasoning functions, ONA distinctly separates sensorimotor inference from semantic inference \cite{hammer2020opennars}. This division allows for specialized handling of different types of reasoning tasks—sensorimotor inference can manage real-time, action-oriented processes, while semantic inference deals with abstract, knowledge-based reasoning. This separation helps to optimize processing efficiency and reduces the computational complexity involved in handling diverse reasoning tasks simultaneously.

\item \textbf{Resource Management:} ONA places a strong emphasis on managing computational resources effectively, adhering to the Assumption of Insufficient Knowledge and Resources (AIKR). It is designed to operate within strict memory and processing constraints, employing mechanisms like priority-based forgetting and constant-time inference cycles. These features ensure that ONA can function continuously in resource-limited settings by efficiently managing its cognitive load and memory usage.

\item \textbf{Advanced Data Structures and Memory Management:} ONA utilizes a sophisticated system of data structures that include events, concepts, implications, and a priority queue system for managing these elements. This setup facilitates more refined control over memory and processing, prioritizing elements that are most relevant to the system's current goals and tasks. It also helps in maintaining the system’s performance by managing the complexity and volume of information it handles.

\item \textbf{Practical Application Focus:} The architectural and control changes in ONA are driven by a focus on practical application needs, which demand reliability and adaptability. ONA is tailored to function effectively in real-world settings that require autonomous decision-making and adaptation to changing environments, making it more applicable and robust than its predecessors for tasks in complex, dynamic scenarios \cite{hammer2020opennars}.
\end{enumerate}

These enhancements position ONA as a capable and versatile system within the NARS family, designed to meet the challenges of practical deployment in varied operational contexts.

For this study, all examples are using a version of ONA compiled with the parameter \verb!SEMANTIC_INFERENCE_NAL_LEVEL! was set to \verb!0!, which means that only NAL layers 6–8 were enabled. This means that the system could only do sensorimotor inference (procedural and temporal reasoning), but no semantic inference (declarative reasoning). In addition to this functionality, a change regarding functional equivalence was implemented as described below.

\section{Functional equivalence with ONA}

For this study, ONA was updated to include a new functionality to enable functional equivalence. One of the simplest cases, where this functionality is demonstrated, is regarding two three-term contingencies, as in the following example.

\begin{verbatim}
A. :|:
^op1. :|:
G. :|:

100

B. :|:
^op1. :|:
G. :|:
\end{verbatim}

With the above example, a derived functional equivalence would be established between $A$ and $B$, since they are both preconditions of the same operation \verb!^op1! leading to the same goal $G$. Importantly, to restrict the introduction of such higher-order statements, this functionality is only activated when operations involved in the contingencies are executed. This allows ONA to avoid a combinatorial explosion caused by an unrestricted use of functional equivalence.

An example of functional equivalence with ONA follows. ONA was set up with the following parameters:

\begin{verbatim}
*babblingops=2
*motorbabbling=0.9
*setopname 1 ^say
*setopname 2 ^select
*volume=100
\end{verbatim}

The following code demonstrates the training of two three-term contingencies:

\begin{verbatim}
<(left * red) --> (loc * color)>. :|:
<({SELF} * RED) --> ^say>. :|:
G. :|:

<(left * r-e-d) --> (loc * ocr)>. :|:
<({SELF} * RED) --> ^say>. :|:
G. :|:
\end{verbatim}

In this example, the first line represents a situation where something of red color is presented in a left location, followed by the system saying \textit{“RED”}, and a consequence $G$. In the fourth line, a written word \texttt{r-e-d} (for example detected by an OCR component of the system) is detected in the left location, followed by a similar operation and consequence as above. 

After acting on both these contingencies, to trigger the formation of functional equivalence, ONA would derive the following:

\begin{verbatim}
<<(left * red) --> (loc * color)> <=> <(left * r-e-d) --> 
 (loc * ocr)>>

<<($1 * r-e-d) --> (loc * ocr)> <=> <($1 * red) --> 
 (loc * color)>>
\end{verbatim}

After having derived this, ONA could deal with the following situation:

\begin{verbatim}
<(up * red) --> (loc * color)>. :|:
<({SELF} * up) --> ^select>. :|:
G. :|:

// Derived:
<(<(#1 * red) --> (loc * color)> &/ <({SELF} * #1) --> ^select>) 
 =/> G>.

100

<(up * r-e-d) --> (loc * ocr)>. :|:
G! :|:

// Executed: <({SELF} * up) --> ^select>

100

<(down * r-e-d) --> (loc * ocr)>. :|:
G! :|:

// Executed: <({SELF} * down) --> ^select>
\end{verbatim}

In the above example, ONA first learns that if selecting “up” when something red is shown in the upper location, a consequence G will happen. A more general contingency (with variables being introduced) is derived. Later, the word \texttt{r-e-d} is presented in an upper position. Importantly, despite that the system has never learned a contingency involving the word \texttt{r-e-d}, it can execute the appropriate actions by substituting the word \texttt{r-e-d} for the red color, as learned by the functional equivalence in the beginning of the example. This demonstrates a unique form of transfer learning with ONA that is enabled by functional equivalence.

\section{Extended example of reading and cross-modal transfer}

\begin{figure}
\centering
\includegraphics[scale=0.3]{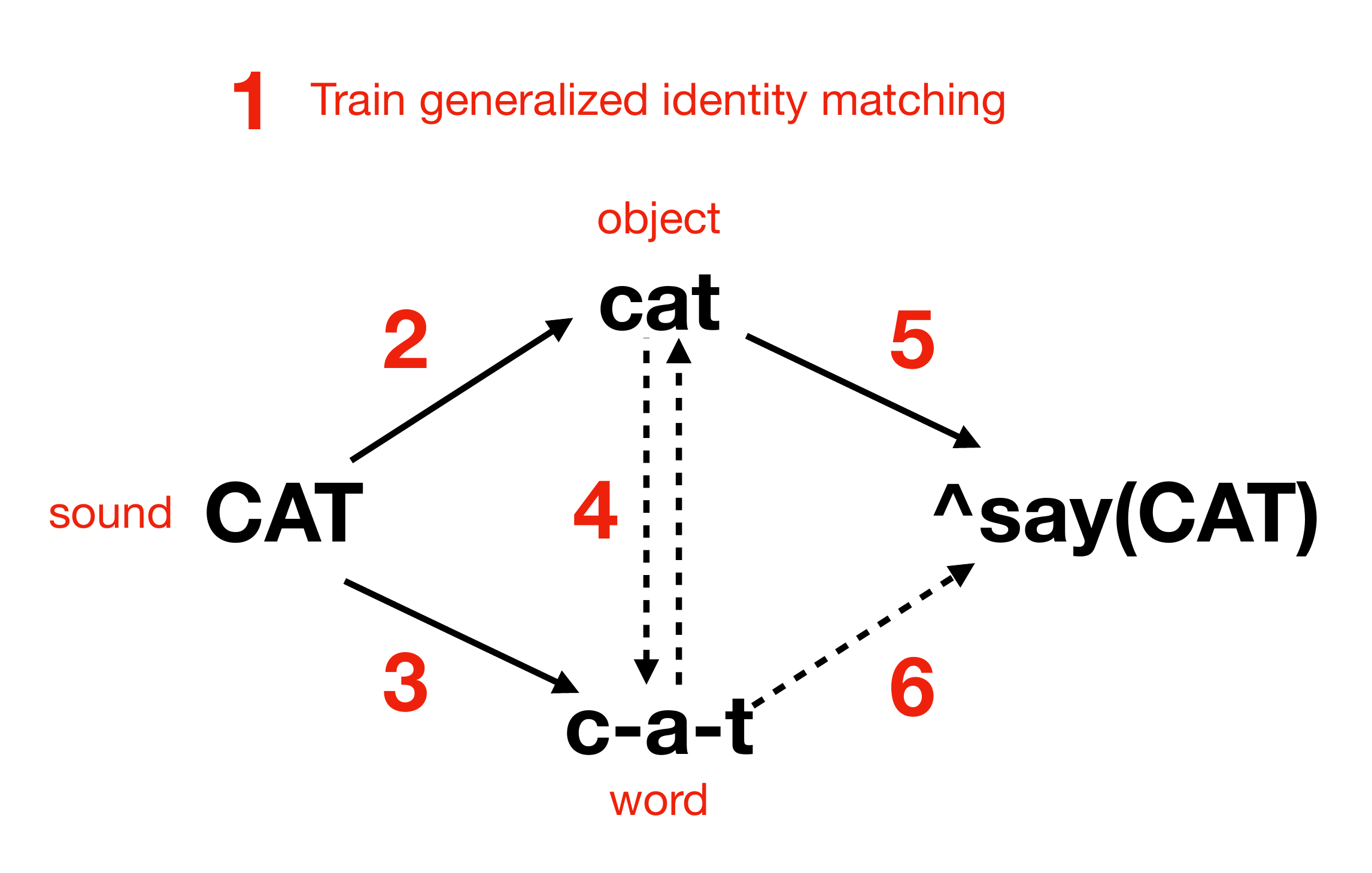}
\caption{
The six step involved in the example.
} 
\label{fig_cat}
\end{figure}

To demonstrate further capabilities enabled by functional equivalence, the following example is used. In the example, an ONA system could learn to “read” and to transfer knowledge across modalities (objects to words, and words to objects), tasks that arguably are fundamental for advanced cognition and general artificial intelligence. The experimental setup is inspired by a 1973 study by Sidman and Cresson \cite{sidman1973reading}. Figure \ref{fig_cat} demonstrates the steps from that particular study. Two young boys (17 and 18 years old) with severe functional disabilities were first trained to match printed words to each other (visual discrimination or identity matching; Step 1 in Figure \ref{fig_cat}), and dictated words to their corresponding pictures (auditory comprehension; Step 2). After that training, they were still unable to match the printed words to their pictures (reading comprehension; Step 4) or read the printed words orally (Step 6). They were next taught to match the dictated to the printed words (Step 3) and were after this training able to demonstrate comprehension (Step 4). After training to name the pictures (Step 5), they were also able to “read” the words orally (Step 6). 

This demonstrated that the learned equivalences of dictated words to pictures and to printed words transferred to the purely visual equivalence of printed words to pictures. 

To demonstrate the above example, an ONA system could be set up in the following way:

\begin{verbatim}
*babblingops=3
*motorbabbling=0.9
*setopname 1 ^say
*setopname 2 ^select
*setopname 3 ^match
*volume=100
\end{verbatim}

\subsubsection{1. Teach word-word matching (Identity matching)}

First, the system would need to be trained in generalized identity matching, as we have demonstrated in a previous publication \cite{johansson2023generalized}. 

\begin{verbatim}
<(sample * a-x-e) --> (loc * ocr)>. :|:
<(left * a-x-e) --> (loc * ocr)>. :|:
<(right * h-a-t) --> (loc * ocr)>. :|:
<({SELF} * (sample * left)) --> ^match>. :|:
G. :|:

// Derived:
<(<($1 * #1) --> (loc * ocr)> &/ <($2 * #1) --> (loc * ocr)>) =/>
 <({SELF} * ($1 * $2)) --> ^match>>.	
\end{verbatim}

\subsubsection{2. Teach sound-object matching (Auditory comprehension)}

Then, the system is trained to match a particular sound (e.g., the spoken word \textit{“CAT”}, to an object (e.g., a cat object identified by an image classifier, like the YOLO model). The matching is indicated by selecting the left or right, depending on the location of the object.

\begin{verbatim}
<CAT --> sound>. :|:
<(left * cat) --> (loc * yolo)>. :|:
<(right * dog) --> (loc * yolo)>. :|:
<({SELF} * left) --> ^select>. :|:
G. :|:

// Derived:
<((<CAT --> sound> &/ <(#1 * cat) --> (loc * yolo)>) &/ 
 <({SELF} * #1) --> ^select>) =/> G>.
\end{verbatim}

\subsubsection{3. Teach sound-word matching (Auditory receptive reading)}

A similar training as in Step 2 would be used to train to pick a certain written word, in relation to a particular sound.

\begin{verbatim}
<CAT --> sound>. :|:
<(left * c-a-t) --> (loc * ocr)>. :|:
<(right * d-o-g) --> (loc * ocr)>. :|:
<({SELF} * left) --> ^select>. :|:
G. :|:

// Derived:
<((<CAT --> sound> &/ <(#1 * c-a-t) --> (loc * ocr)>) &/ 
 <({SELF} * #1) --> ^select>) =/> G>.
\end{verbatim}

The derivations from Step 2 and 3 would then be:

\begin{verbatim}
<((<CAT --> sound> &/ <(left * cat) --> (loc * yolo)>) &/ 
 <({SELF} * left) --> ^select>) =/> G>.

<((<CAT --> sound> &/ <(left * c-a-t) --> (loc * ocr)>) &/ 
 <({SELF} * left) --> ^select>) =/> G>.
\end{verbatim}

A functional equivalence could then be derived:

\begin{verbatim}
<(<CAT --> sound> &/ <(left * cat) --> (loc * yolo)>) <=>
 (<CAT --> sound> &/ <(left * c-a-t) --> (loc * ocr)>)>
\end{verbatim}

Since the left-hand side of both sides of the equivalence are the same, the following could be derived:

\begin{verbatim}
<<(left * cat) --> (loc * yolo)> <=> 
 <(left * c-a-t) --> (loc * ocr)>>

<<($1 * cat) --> (loc * yolo)> <=> 
 <($1 * c-a-t) --> (loc * ocr)>>
\end{verbatim}

\subsubsection{4. Test object-word matching (Reading comprehension)}

The following situation would be presented, that the system never has experienced before:

\begin{verbatim}
<(sample * cat) --> (loc * yolo)>. :|:
<(left * c-a-t) --> (loc * ocr)>. :|:
<(right * d-o-g) --> (loc * ocr)>. :|:
G! :|:
\end{verbatim}

With the following derivation from Step 3:

\begin{verbatim}
<<($1 * cat) --> (loc * yolo)> <=> <($1 * c-a-t) --> (loc * ocr)>>
\end{verbatim}

The situation would be transformed to the following:

\begin{verbatim}
<(sample * c-a-t) --> (loc * ocr)>. :|: // Substituted
<(left * c-a-t) --> (loc * ocr)>. :|:
<(right * d-o-g) --> (loc * ocr)>. :|:
G! :|:

// Executed (using identity matching from Step 1):
<({SELF} * (sample * left)) --> ^match>
\end{verbatim}

\subsubsection{5. Teach object naming}

The system would then be trained to say a particular word in relation to objects:

\begin{verbatim}
<(up * cat) --> (loc * yolo)>. :|:
<({SELF} * CAT) --> ^say>. :|:
G. :|:	
\end{verbatim}

\subsubsection{6. Test word naming (Oral reading)}

A situation like the following (that has not been experienced previously) could then be presented:

\begin{verbatim}
<(up * c-a-t) --> (loc * ocr)>. :|:
G! :|:
\end{verbatim}

With the following, as derived in Step 3:

\begin{verbatim}
<<($1 * cat) --> (loc * yolo)> <=> <($1 * c-a-t) --> (loc * ocr)>>	
\end{verbatim}

The situation can then be transformed to the following:

\begin{verbatim}
<(up * cat) --> (loc * yolo)>. :|: // Substituted
G! :|:

// Executed:
<({SELF} * CAT) --> ^say>.
\end{verbatim}

\subsubsection{Summary}

This example demonstrated how a NARS system could be trained in semantic relations between sounds, words, and objects, using derivations based on functional equivalence.

\section{Discussion}

In this paper, we introduced the concept of functional equivalence, defined as a derived equivalence, based on a shared similar function in contingencies. This seems fundamental in understanding how cognitive systems, such as NARS, can derive meaning and knowledge from their interactions with the environment. The principle of functional equivalence allows for the abstraction of concepts beyond their immediate instances, enabling cognitive systems to apply learned knowledge in novel situations. The examples in this paper arguably demonstrate a key aspect of cognitive flexibility.

Moreover, the methodology presented in this paper, inspired by studies carried out with individuals with severe functional disabilities, highlights the potential for training AGI systems to be capable of complex linguistic tasks, by a systematic step-wise approach that focuses on incrementally increasing the complexity of tasks. This pathway for teaching machines to understand and generate “building blocks” of human-like language could potentially inspire how several different AGI systems are developed and implemented.

Future research could explore the application of functional equivalence in more complex scenarios, including dynamic environments and social interactions, for example with NARS on robotics. Additionally, investigating the limits of functional equivalence in cognitive systems would provide valuable insights into both the potential and the challenges of AGI in achieving human-like understanding and reasoning abilities.

In conclusion, the concept of functional equivalence seems to offer a key capability in AGI systems, that for NARS, is realized by equivalence as suggested by Non-Axiomatic Logic \cite{wang2013nalbook}. By highlighting the mechanisms through which systems like NARS can derive meaning from their interactions, this research contributes to the broader goal of creating human-level AGI. Further exploration in this direction seems warranted.

\begin{credits}

\subsubsection{\discintname}
The authors have no competing interests to declare that are
relevant to the content of this article.

\end{credits}
%
%
%
%

\bibliographystyle{splncs04}
\bibliography{RJ}

\begin{thebibliography}{1}
\providecommand{\url}[1]{\texttt{#1}}
\providecommand{\urlprefix}{URL }
\providecommand{\doi}[1]{https://doi.org/#1}

\bibitem{hammer2020opennars}
Hammer, P., Lofthouse, T.: ‘opennars for applications’: architecture and control. In: International Conference on Artificial General Intelligence. pp. 193--204. Springer (2020)

\bibitem{johanssonlofthouse2023stimulus}
Johansson, R., Lofthouse, T.: Stimulus equivalence in nars. In: International Conference on Artificial General Intelligence. pp. 158--166. Springer (2023)

\bibitem{johansson2023generalized}
Johansson, R., Lofthouse, T., Hammer, P.: Generalized identity matching in nars. In: Artificial General Intelligence: 15th International Conference, AGI 2022, Seattle, WA, USA, August 19--22, 2022, Proceedings. pp. 243--249. Springer (2023)

\bibitem{lofthouse2019alann}
Lofthouse, T.: Alann: An event driven control mechanism for a non-axiomatic reasoning system (nars). In: NARS2019 workshop at AGI (2019)

\bibitem{schusterman1998functional}
Schusterman, R.J., Kastak, D.: Functional equivalence in a california sea lion: Relevance to animal social and communicative interactions. Animal Behaviour  \textbf{55}(5),  1087--1095 (1998)

\bibitem{sidman1973reading}
Sidman, M., Cresson, O.: Reading and crossmodal transfer of stimulus equivalences in severe retardation. American Journal of Mental Deficiency  (1973)

\bibitem{wang2013nalbook}
Wang, P.: Non-axiomatic logic: A model of intelligent reasoning. World Scientific (2013)

\end{thebibliography}






\end{document}